\documentclass[conference]{IEEEtran}
\IEEEoverridecommandlockouts

\usepackage{cite}
\usepackage{amsmath,amssymb,amsfonts}
\usepackage{algorithmic}
\usepackage{graphicx}
\usepackage{textcomp}
\usepackage{xcolor}
\usepackage{comment}
\usepackage{url}
\usepackage{booktabs}
\usepackage{algorithmic}
\usepackage{algorithm}
\usepackage{multirow}
    
\usepackage[pscoord]{eso-pic} 

\begin{document}

\title{WEST GCN-LSTM: Weighted Stacked Spatio-Temporal Graph Neural Networks for Regional Traffic Forecasting}

 \author{Theodoros Theodoropoulos,
        Angelos-Christos Maroudis,~Antonios Makris, and~Konstantinos Tserpes\\
\IEEEauthorblockA{\textit{Department of Informatics and Telematics, Harokopio University of Athens, Greece}
}
\IEEEauthorblockA{\textit{School of Electrical and Computer Engineering, National Technical University of Athens, Greece}
}
}


\maketitle

\begin{abstract}
Regional traffic forecasting is a critical challenge in urban mobility, with applications to various fields such as the Internet of Everything. In recent years, spatio-temporal graph neural networks have achieved state-of-the-art results in the context of numerous traffic forecasting challenges. This work aims at expanding upon the conventional spatio-temporal graph neural network architectures in a manner that may facilitate the inclusion of information regarding the examined regions, as well as the populations that traverse them, in order to establish a more efficient prediction model. The end-product of this scientific endeavour is a novel spatio-temporal graph neural network architecture that is referred to as WEST (WEighted STacked) GCN-LSTM. Furthermore, the inclusion of the aforementioned information is conducted via the use of two novel dedicated algorithms that are referred to as the Shared Borders Policy and the Adjustable Hops Policy. Through information fusion and distillation, the proposed solution manages to significantly outperform its competitors in the frame of an experimental evaluation that consists of 19 forecasting models, across several datasets. Finally, an additional ablation study determined that each of the components of the proposed solution contributes towards enhancing its overall performance.
\end{abstract}


\section{Introduction}
Regional traffic forecasting is an emerging challenge in the domain of urban mobility that holds significance in various fields such as smart cities \cite{verma2021extracting}, edge computing \cite{shi2016edge}, the Internet of Things \cite{zanella2014internet}, wireless networks \cite{schindelhauer2006mobility}, personalised recommender systems \cite{singh2021recommender}, epidemiology modeling \cite{ilin2021public}, and many more. However, the significance of regional traffic forecasting escalates within the Internet of Everything (IoE) paradigm \cite{snyder2017internet}, which is characterized by an intricate web of relationships among people, things, data, and processes. Regional traffic forecasting refers to the process of predicting future traffic conditions across diverse geographic areas, characterized by grid-based or non-uniform partitioning, and over multiple periods of time, which may span from a couple of minutes to several hours. This process is conducted via the use of dedicated forecasting models. 

Despite its numerous beneficial applications in the frame of the aforementioned fields, regional traffic forecasting poses a complex problem, as it requires accurately forecasting traffic conditions across different areas over an extended time period. This complexity arises from the intricate and interlinked two-fold nature of traffic systems that manifest both spatial temporal characteristics \cite{alessandretti2017multi}. Spatially, traffic conditions in one region can be influenced by events occurring in neighboring or distant areas, requiring an understanding of the spatial dependencies between different regions. In other words, traffic conditions in one area can have a cascading effect on neighboring regions, making it imperative to accurately model and account for the interdependencies among regions. Temporally, traffic patterns undergo dynamic changes influenced on the basis of time cycles of varying lengths \cite{barbosa2018human}, demanding models that capture both short-term fluctuations and long-term trends. 

Thus, any attempt at constructing regional traffic forecasting models should be designed in a manner that incorporates the use of information regarding the topology of the various regions and the  populations that traverse them. Thankfully, such information can become readily available via the use of technologies such as advanced traffic sensor networks \cite{po2019sensors}, and integrated geographic information systems \cite{cao2023understanding}. These technologies are capable of continuously monitoring, documenting, and archiving spatial and time information across multiple instances. By combining advanced modeling techniques, real-time data streams, and domain-specific knowledge, it is possible to create robust and accurate forecasting systems capable of handling the intricacies of regional traffic \cite{kashinath2021review}.

In recent years, spatio-temporal graph neural networks, such as GCN-LSTM \cite{liu2020method}, have demonstrated state-of-the-art performance in a wide range of traffic forecasting problems \cite{9511330}, primarily due to their intrinsic nature of effectively incorporating contextual information. However, only a small portion of these endeavours focuses on regional traffic forecasting. Furthermore, the use of spatio-temporal graph neural networks for regional traffic forecasting has been quite limited in the sense that all prior attempts focus on capturing either the spatial or the temporal aspects of this problem. 

This work aims at expanding upon the GCN-LSTM architecture in a manner that may facilitate the incorporation of information regarding the various populations (temporal aspect), as well as the regions that they traverse (spatial aspect), in order to establish more refined and accurate prediction models, through information fusion and distillation. The result of this scientific endeavour is a novel spatio-temporal graph neural network architecture that leverages weighted stacked graph convolution. This architecture is referred to as WEST (WEighted STacked) GCN-LSTM. Furthermore, the incorporation of the aforementioned information is conducted via the use of two novel algorithms that are referred to as the Shared Borders Policy and the Adjustable Hops Policy. This paper is dedicated to analyzing the proposed architecture and algorithms in great detail, and to evaluating the efficiency of the proposed solution that consists of them.

More specifically, the rest of the paper is structured in the following manner: Section \ref{Sec:related} explores the corresponding scientific literature that is based on the use of various traffic forecasting models. Section \ref{Sec:formulation} establishes the problem formulation that shall be used throughout this paper. Section \ref{Sec:proposed} showcases the proposed solution. Section \ref{Sec:experiment} describes the experimental process undertaken to evaluate the efficiency of the proposed solution. Finally, Section \ref{Sec:Conclusion} summarizes the merits and findings of this work, and proposes potential future research directions.

\section{Literature Review}
\label{Sec:related}
Depending on the examined regional traffic forecasting scenario, the characteristics of the moving entities, as well as the regions that they traverse may vary significantly. Nevertheless, regardless of which type of scenario is being considered, these two aspects constitute the cornerstone of regional traffic forecasting. As such, any attempt at conducting regional traffic forecasting should be carried out within a spatio-temporal framework that is capable of encapsulating the dynamic nature of the moving entities that is bound to manifest. 

This view of traffic evolution as a dynamic system led earlier attempts at traffic forecasting to consider the use of Recurrent Neural Networks (RNNs) \cite{van2002freeway}. While RNNs have been particularly effective for modeling sequential data exhibiting dynamic behavior, their efficacy wanes in storing prolonged information, attributed to the vanishing / exploding gradient phenomenon that appears in long sequence learning \cite{hochreiter2001gradient}. To surmount this limitation, Long Short-Term Memory (LSTM) networks were used in the context of regional traffic forecasting \cite{ma2015long}. LSTMs excel in capturing temporal intricacies and long-term dependencies. However, while these models proficiently handle sequential data and unveil temporal patterns within mobility tasks, their prowess in the temporal domain stands in contrast to their limitations in comprehensively encapsulating the spatial aspects of the problem.

Encoder-decoders (EDs) are composite Deep Learning (DL) architectures that are capable of mitigating this limitation. They are designed to handle variable-length input and output sequences, making it ideal for sequence-to-sequence predictions due to their structure. The encoder converts variable-length input into a fixed-shape state, and the decoder, using the encoder's states, generates the output based on gathered information. While the role of the encoder is encapsulate the spatial underlying dependencies, the decoder aims at capturing the various temporal patterns and thus is usually based on some form of RNN models. Notable encoder-decoder architectures that have been examined in the frame of traffic forecasting scenarios include the LSTM ED \cite{du2019lstm}, THE BD-LSTM ED \cite{nadeeshan2021multi}, the CNN-LSTM \cite{cao2020cnn}, the Hybrid LSTM ED \cite{theodoropoulos2021encoder}, and the Hybrid LSTM ATT ED \cite{violos2022self}. 

More specifically, the findings that derived while authoring the latter of these works, motivated us to focus our efforts towards establishing a more advanced solution for regional traffic forecasting. This work showcased that ED architectures, when leveraged for multi-step regional traffic forecasting in the context of a single region, manage to outperform their competitors. However, the situation changes drastically when simultaneously exploring multiple regions in a multi-step manner, due to the dramatic increase in the complexity of the input and output sequences. During such multi-regional scenarios, ED architectures seem to lose their clear competitive advantage against other approaches, such as the ones that are based on Linear Regression (LR) \cite{su2012linear}, and Machine Learning (ML) \cite{jordan2015machine} paradigms, depending on the underlying characteristics of the dynamic systems that derive from the corresponding regional traffic scenario. Significant increases in the complexity of the input and output sequences seem to construct a performance plateau that disproportionally affects each forecasting approach. This observation showcased the need to introduce regional traffic forecasting mechanisms whose efficiency is not jeopardized by increases in the underlying complexity of the problem, but instead through information fusion and refinement manage to consistently rise above the aforementioned plateau, regardless of the underlying system dynamics.

In recent times, the pursuit of effective methodologies to address challenges inherent in processing data originating from non-Euclidean domains has garnered significant attention. At the forefront of these endeavors lie Graph Neural Networks (GNNs) \cite{1555942}, renowned for their adeptness in resolving problems that are intricately intertwined with spatial aspects \cite{wu2020comprehensive}. This prowess emanates from their inherent capacity to harness and exploit the spatial attributes of data pertaining to a given problem. Over time, foundational GNN architectures have undergone extensions aimed at augmenting their inherent attributes, thus resulting in several variations. The most notable of these variations are recurrent graph neural networks \cite{dai2018learning}, graph convolutional networks (GCN) \cite{zhang2019graph}, graph autoencoders \cite{hou2022graphmae}, and spatio-temporal graph neural networks \cite{bui2022spatial}. Spatio-temporal graph neural networks have been proven quite successful in the field of traffic forecasting \cite{liu2024rt}, since their architecture enables them to simultaneously capture spatial and temporal dependencies. This can be achieved by leveraging graph convolutions to model spatial dependencies alongside RNNs to encapsulate temporal dependencies \cite{zhao2019t}, in a manner that is aligned with the ED paradigm. 

The modus operandi of GNNs is based on iteratively aggregating information from the neighbors and updating the representations of nodes. The neighbors of each node are dictated by a dedicated adjacency matrix. In most cases, the node representation is updated on the basis of the direct neighbors of the node, which are called 1-hop neighbors. However, there have been works \cite{abu2019mixhop, nikolentzos2020k} that advocate for the extension of information aggregation to K-hop format, in order to enhance the model's expressive power \cite{feng2022powerful}. In the context of spatio-temporal graph neural networks there has been only a single instance \cite{zhao2022k} that the incorporation of K-hop information aggregation was examined. However, in the context of the aforementioned work, $K$ was regarded as a parameter that is tuned by trial-and-error. In our work however, $K$ is calculated using information regarding the moving speed of the populations and the topology of the regions. This design choice enables the proposed solution to encapsulate the temporal aspects of regional traffic forecasting, in regards to the moving speed of the various populations.

Despite the fact that there have been numerous works that examine the use of GNNs in the context of traffic forecasting \cite{jiang2023graph}, only a rather small fraction of them focus on the use of graph neural networks for regional traffic forecasting \cite{jiang2022graph}, the vast majority of which can be categorised based on the way they incorporate the graphs' adjacency matrices. Most of these works \cite{chen2020dynamic, peng2020spatial, zhou2020exploiting, yang2022region, wang2022multi, li2022multigraph} choose to construct the adjacency matrices based on various distance-related metrics (in most cases the distance between the centers of the regions). Other works \cite{qiu2020topological, sun2020predicting, wang2020multi} propose the construction of the adjacency matrices on the basis of traffic pattern similarity matrices. Finally, the last category includes solutions \cite{wang1804graph, shi2020predicting, yeghikyan2020learning} that leverage binary adjacency matrices based on whether or not the involved nodes are neighboring. In our work we expand upon the latter category by proposing the use of a weighted adjacency matrix in order to encapsulate the lengths of borders that are being shared between the various regions. This design choice enables the proposed solution to encapsulate the spatial aspects of regional traffic forecasting, in regards to the topology of the involved regions in a more refined manner. Furthermore, all of the aforementioned GNN-based solutions for regional traffic forecasting are designed to focus on either the spatial (distance of centers, neighboring status) or the temporal (traffic) aspects of this problem. However, in our work we propose a solution that is capable of capturing both the spatial and the temporal aspects of regional traffic forecasting in an optimal manner.

Our work aims to introduce an advanced forecasting model that through information refinement \& fusion, is capable of producing more accurate regional traffic predictions. Towards achieving this goal, we extend spatio-temporal graph neural networks in a manner that is aligned with the regional traffic forecasting paradigm. In order to do so, we propose a novel spatio-temporal graph neural networks architecture that incorporates weighted stacked convolutions. Weighted stacked convolutions require the calculation of the weighted adjacency matrix and the number of graph convolution layers $K$. To that end, we also propose two novel policies that leverage information regarding speed of the populations and the topography of the regions that they traverse, to calculate the adjacency matrix and $K$. 

\section{Problem Formulation}
\label{Sec:formulation}

This paper presents an evaluation of models designed to forecast population flow estimates for various regions across multiple future time periods. We refer to this challenge as multi-step regional traffic forecasting. A Region refers to a specific area within a larger geographic or urban context that is of particular interest for a certain purpose or analysis. The Regions are represented by the set $N = \{n_1,n_2..,n_n\}$, where $n_n$ indicates the $n^{th}$ Region, where $1\leq n \leq |N|$. Each $Region_n$ is characterized by its Borders which are represented by the set $B_n = \{b_1,b_2..,b_b\}$, where $b_b$ indicates the $b^{th}$ Border, where $1\leq b \leq B$, and by its $Center_n$ which is represented by a 2-tuple that corresponds to its $x$ \& $y$ coordinates. The average distance between the various $Center_n$ 2-tuples is denoted by $D$. Furthermore, each Border is regarded as a line segment, which is characterized by a 4-tuple that corresponds to the $x$ \& $y$ coordinates of the two endpoints of the line segment. The various types of Populations are represented by the set $P = \{p_1,p_2..,p_p\}$, where $p_p$ indicates the $p^{th}$ Population, where $1\leq p \leq |P|$. The distinction between Population types is made based on their ability to traverse the area, and thus each of these Populations is characterized by an average moving speed $Speed_p$. The notations used in the context of this work are showcased in Table \ref{tab0}.

In time-series analysis, the multi-step formulation involves predicting future values of a time series by forecasting multiple time-steps ahead. This method differs from the single-step approach, which only estimates the next point in time. Furthermore, the multivariate formulation in time series analysis involves creating a model for target variables that relies on multiple predictor variables. These predictor variables are interdependent and display temporal dependencies over time, and they may be impacted by exogenous inputs and noise. This method can be represented mathematically as a system of equations, where the target variable and predictor variables are modeled as stochastic processes that vary with time.

In the context of the present challenge, the output vector's dimensional space is denoted by $R^{|N|*U}$, wherein $|N|$ represents the number of Regions for which we intend to predict traffic at a given time point $t$, and $U$ denotes the number of future steps over which we aim to make these projections. Similarly, we define the input vector's dimensional space as $R^{|N^{'}|*U^{'}}$, where $|N^{'}|$ corresponds to number of the Regions whose population variations exhibit a reliance on those of $N$, and $U^{'}$ signifies the number of preceding time steps that contribute to the retrospective observation window (look-back window). It is pertinent to note that in our modeling, the value of $|N^{'}|$ is equal to that of $|N|$. 

In further elaboration, we focus on a particular time point $t_{i}$ and consider the input vector $X$ as follows: 

\begin{equation} \label{eq:{Input_Vector_X}}
        X = \{x_{i-U^{'}+1}, ..., x_{i-l'}, ..., x_{i}\}, l' \in U',
\end{equation}

,wherein $x_{i-l'} = \{Region_{t_{i-l'}}^{1},Region_{t_{i-l'}}^{2},...,Region_{t_{i-l'}}^{n^{'}} \}$ represents the Population of each Region $n \in N^{'}$ at the time $t_{i-l'}$. In a similar manner, we model the output vector $Y$, which is characterized as follows:

\begin{equation} \label{eq:{Output_Vector_Y}}
        Y = \{y_{i+1}, ..., y_{i+l}, ..., y_{U}\}, l \in U,
\end{equation}

,wherein $y_{i+l} = \{Region_{t_{i+l}}^{1},Region_{t_{i+l}}^{2},...,Region_{t_{i+l}}^{n} \}$ represents the population of each Region $n \in N$ at the time $t_{i+l}$.

Since our work aims to expand upon spatio-temporal graph neural networks, it is of vital importance to convert the prior problem formulation to graph format. There have been numerous variations in terms of types of graphs. Arguably, the most significant distinction among these variations lies in whether the considered graph structures are static or dynamic. Dynamic graphs can be classified into Discrete-Time Dynamic Graphs (DTDG) \cite{Yu_2018} and Continuous-Time Dynamic Graphs (CTDG) \cite{https://doi.org/10.48550/arxiv.2006.10637}. The authors of this work have chosen the DTDG approach in order to formulate regional traffic in a dynamic manner. According to the DTDG paradigm, a dynamic graph is defined as a sequence of snapshots of a static graph. Each one of the snapshots corresponds to a specific time-step $t$, the duration of which is referred to as $t_{window}$. These snapshots construct a temporal continuum that enables the emergence of temporal patterns and phenomena. Furthermore, each one of the aforementioned static graphs consists of multiple nodes and edges that encapsulate the underlying spatial relations. In the context of this work, each graph corresponds to an area in two-dimensional space that is divided in $N$ Regions that are being traversed by $|P|$ Populations at each time-step $t$. 

Given an undirected graph $\overline G$ that consists of $|N|$ nodes and $E$ edges. The nodes of the graph correspond to the Regions, and the edges of the graph correspond to how likely it is for a member of a Population to move from one Region to another, within the time-frame of a singular time-step $t$. This graph can be described by the following two matrices:

\begin{itemize}
    \item A weighted \textbf{Adjacency Matrix} $\mathbf{A} \in \mathbb{R}^{|N| \times |N|}$ that incorporates edge weights $w{ij}$.
    \item A \textbf{Feature Matrix} $\mathbf{Z} \in \mathbb{R}^{|N| \times F}$, where $F$ corresponds to the dimension of each Feature Vector. 
\end{itemize}

The Feature Matrix can be viewed as the total of the Feature Vectors. Each one of the $|N|$ rows of the Feature Matrix corresponds to a Feature Vector that describes node-level features. In the context of this work, each node (Region) is described by a Feature Vector, whose dimension is equal to $U^{'}$ which are the traffic values recorded at the corresponding Region during the last $U^{'}$ time-steps, and thus $F$ = $U^{'}$. Furthermore, each of the $U^{'}$ columns of the Feature Matrix $Z$ corresponds to a different time-step $t$ of the input sequence. This formulation enables instances of the Feature Matrix to be modeled as time-series data. The Adjacency Matrix $A$ is static and thus, remains unchanged throughout the various time-steps, since it models the statistical possibility of moving from one Region to another. On the other hand, the Feature Matrix $Z$ is dynamic and is different for each time-step $t$. Subsequently, snapshots of the Feature Matrix $Z$ are conceptually equivalent to the aforementioned input vector $X$ $\in \mathbb{R}^{|N| \times U^{'}}$.

\begin{table}[]
\caption{Notations used in this paper.\label{tab0}}
\centering
\begin{tabular}{@{}ll@{}}
\toprule
\textbf{Notations}  & \textbf{Descriptions} \\ \midrule
$\mathbb{R}^{n}$          & n-dimensional euclidean space \\
$N$                       & set of Regions \\
$Border_{n}$              & set of $Region_n$'s Borders \\
$Center_{n}$              & Center of $Region_n$ \\
$D$                       & the average distance between the Centers \\
$P$                       & set of Populations \\
$Speed_{p}$               & average speed of $Population_p$ \\
$U$                       & number of input time-steps \\
$U'$                      & number of prediction time-steps \\
$t_{window}$              & duration of each time-step \\
$F$                       & dimension of Feature Vector \\
$Z$                       & Feature Matrix \\
$D$                       & Average distance between two Region Centers\\
$\overline G$             & Undirected Graph   \\
$A$                       & Adjacency Matrix   \\
$I$                       & Identity Matrix    \\
$w$                       & Edge Weights    \\
$K$                       & number of stacked GCN layers\\
$W$                       & learnable weight matrix \\
$|X|$                     & the number of elements in a given set X \\
\bottomrule
\end{tabular}
\end{table}

\section{Proposed Solution}
\label{Sec:proposed}

\begin{figure*}[!h]
	\centering
		\includegraphics[scale=0.31]{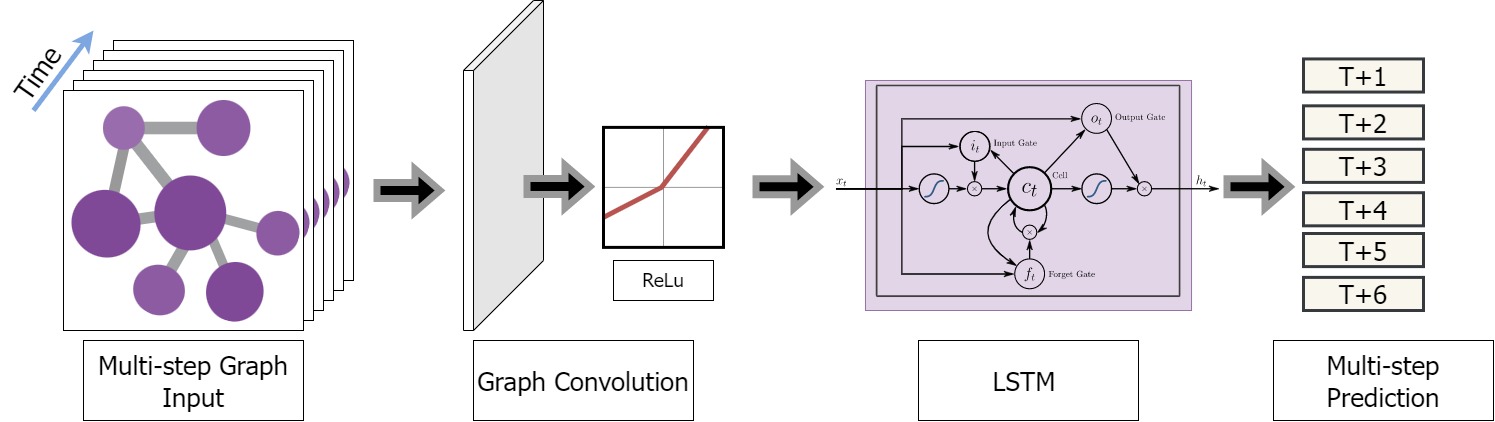}
	\caption{GCN-LSTM architecture.}
	\label{FIG:ARCH}
\end{figure*}

The aim of this work is to expand upon the GCN-LSTM architecture, depicted in Fig. \ref{FIG:ARCH}, in a manner that may enable the incorporation of information regarding the various Populations, as well as the Regions that they traverse, in order to produce more refined and accurate prediction models.

Towards achieving this goal, the proposed solution consists of three components that are closely intertwined with each other. These components are the following ones:
\begin{itemize}
    \item the \textbf{WEST GCN-LSTM}
    \item the \textbf{Shared Borders Policy}
    \item the \textbf{Adjustable Hops Policy}
\end{itemize}

\begin{figure*}[!h]
	\centering
		\includegraphics[scale=0.58]{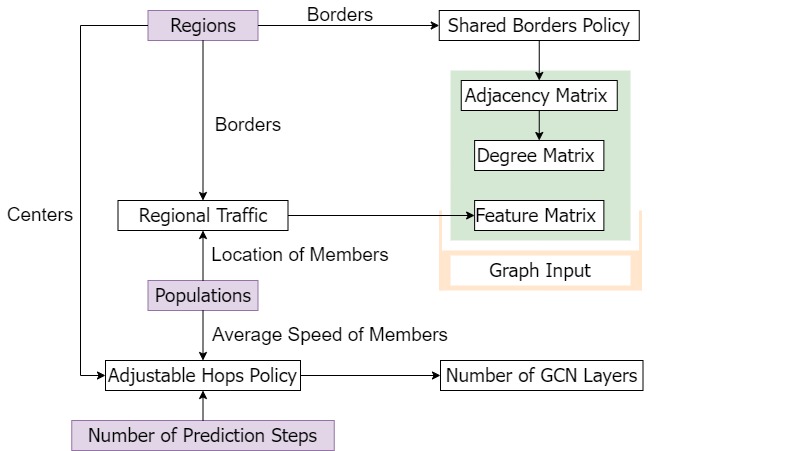}
	\caption{An overview of the proposed solution.}
	\label{FIG:ARCH2}
\end{figure*}

The proposed WEST GCN-LSTM is a novel architectural paradigm that extends the GCN-LSTM architecture by facilitating multiple weighted stacked graph convolution layers, based on a weighted Adjacency Matrix $A$ and the number of stacked graph convolution layers that is denoted by $K$. According to our solution, $A$ and $K$ are calculated using the Shared Borders Policy and the Adjustable Hops Policy, respectively. The Shared Borders Policy is designed to leverage information regarding the Borders of the Regions, while the Adjustable Hops Policy is designed to leverage information regarding the Speed of Populations, the Centers of the Regions, and the number of prediction steps. This section is dedicated to showcasing these three components. An overview of the proposed solution is depicted in Fig. \ref{FIG:ARCH2}.

\subsection{WEST GCN-LSTM}
WEST GCN-LSTM is a spatio-temporal graph neural network that is based on the ED architectural paradigm. As the name suggests, the encoder is based on weighted stacked GCNs and the decoder on LSTMs. In order to establish the foundations of the proposed solution, it is required to briefly delve into the mechanics of the WEST GCN-LSTM model.  

\subsubsection{WEighted STacked (WEST) Graph Convolutional Networks (GCN)}
GCNs leverage the graph structure to aggregate information from neighboring nodes. In this step, each node collects information from its neighbors, including itself, to update its feature representation. This is achieved through a linear transformation and aggregation process. After aggregating neighbor information, a non-linear activation function (ReLU) is applied in order to generate the aggregated representation $h$. It is of paramount importance to note that the number of stacked GCN layers is equal to the number of hops in terms of neighbors that the aggregation process can encompass during each iteration. For instance, in a scenario where a GCN is constructed with just a single convolution layer, nodes can exclusively access their immediate neighbors for aggregating representation data. In the frame of weighted stacked graph convolution, the convolutional layers are influenced by the weighted Adjacency Matrix $A$, while the quantity of stacked graph convolution layers is denoted as $K$. Thus, in the context of weighted stacked graph convolutions, $h$ is calculated in the following manner:

\begin{equation}
h^{(0)} = \left(Z \right) \qquad  \text{for } k = 0
\end{equation}
\begin{equation}
h^{(k)} = \sigma(\hat{D}^{-\frac{1}{2}} \hat{A} \hat{D}^{-\frac{1}{2}} h^{(k-1)} W^{(k-1)}) \qquad  \text{for } k = 1, \ldots, K
\end{equation}

,where $h^{(k)}$ corresponds to the aggregated representation after $k$ convolutional layers, and $h^{(0)} = Z$ corresponds to the Feature Matrix. Furthermore, $\hat{A}$ is equal to $A + I$ and its purpose is to incorporate self-connections, $\hat{D}$ is the diagonal degree matrix of $\hat{A}$, $W$ signifies a dedicated learnable weight matrix, and $\sigma$ represents the ReLU function. 

\subsubsection{Long Short Term Memory (LSTM)}

LSTM networks, similar to their precursor, employ the Hidden State mechanism in order to facilitate the representation of dynamic temporal behaviour. The unique aspect of LSTM networks is their utilization of the Cell State structure. This architecture introduces Cell State manipulation through regulatory mechanisms known as Gates. Each LSTM node encompasses three gate-related elements, all incorporating sigmoid layers to ensure differentiability within the $0-to-1$ range. The sigmoid activation function scales values to facilitate data importance assessment and decision-making regarding retention or omission. Gate structures incorporate two sets of weight matrices, denoted as $W$ and $U$, associated with Hidden State and input, along with additional matrices for Cell State. The input $X_{t}$ corresponds to timestamp $t$. Gates employ these matrices, along with input and prior Hidden State ($hidden_{t-1}$).

The Forget Gate determines which historical information from past timestamps is to be excluded from the Cell State. Its output is computed using Eq. \ref{eq:2}. The Input Gate evaluates the significance of recent input, updating Cell State using Eq. \ref{eq:3}. Cell State calculation employs the $\overline{\text{C}}$ vector, generated as per Eq. \ref{eq:4}, with tanh activation mitigating gradient issues. The Cell State update process is described in Eq. \ref{eq:5}, combining the output of the Forget Gate and the Input Gate with $\overline{\text{C}}$. The Output Gate computes the subsequent hidden state using Eq. \ref{eq:6}. The new Hidden State is calculated according to Eq. \ref{eq:7}. Updated Cell State and Hidden State are then propagated to subsequent LSTM nodes for the next time-step \cite{chung2014empirical}.

\begin{align}
\label{eq:2}   \text{forget}_t = \text{sigmoid}(X_t \cdot W_f + \text{hidden}_{t-1} \cdot U_f) \\
\label{eq:3}   \text{input}_t = \text{sigmoid}(X_t \cdot W_i + \text{hidden}_{t-1} \cdot U_i) \\
\label{eq:4}   \overline{\text{C}} = \text{tanh}(X_t \cdot W_c + \text{hidden}_{t-1} \cdot U_c) \\
\label{eq:5}   C_t = \text{forget}_t \cdot C_{t-1} + \text{input}_t \cdot \overline{\text{C}}_t \\
\label{eq:6}  \text{output}_t = \text{sigmoid}(X_t \cdot W_o + \text{hidden}_{t-1} \cdot U_o) \\
\label{eq:7}  \text{hidden}_t = \text{output}_t \cdot \text{tanh}(C_t)
\end{align}

\subsubsection{WEST GCN-LSTM}

The authors of this work combined weighted stacked GCN and LSTM layers to predict regional traffic. This fusion of spatial and temporal DL layers is referred to as WEST GCN-LSTM. Weighted stacked GCNs constitute the encoder and are designed to extract structural characteristics of the input sequence, producing an aggregated representation. This process is carried out in the following manner:

\begin{equation}
h^{\text{encoder}} = \text{WEST GCN}_{\text{encoder}}(Z, A) \\
\end{equation}

Here, $h^{\text{encoder}}$ represents the aggregated representation after applying weighted stacked graph convolution, $A$ is the weighted Adjacency Matrix of the graph, and $Z$ is the Feature Matrix. This representation is then fed as input to the LSTM part of the model, thus capturing temporal patterns at the graph snapshot level. The LSTM part of the model, acting as a decoder, produces the desired predictions, in the following manner:

\begin{equation}
Y = \text{LSTM}_{\text{decoder}}(h^{\text{encoder}})
\end{equation}

$\text{LSTM}_{\text{decoder}}$ refers to the LSTM network that takes the aggregated representation from the encoder as input in order to produce an output that is then passed through a dense layer in order to generate the multi-step predictions. In the frame of multi-step time-series forecasting, the WEST GCN-LSTM model takes as input a sequence of graph signals, where each signal corresponds to a different time-step and is represented as a graph signal on a fixed graph. The goal is to predict the future values of the time-series based on the graph signals of previous time steps.

\subsection{Shared Borders Policy}
The edges of the graph play an integral role in representing the spatial relations between the Regions by dictating which nodes shall partake in the feature aggregation process. The authors of this paper propose a novel approach by introducing the Shared Borders policy. The Shared Borders policy is based on the reasonable assumption that the greater the lengths of the shared borders are, the more statistically likely it is for a larger percentage of a Population to traverse them. 

According to this approach, the Adjacency Matrix $A$ shall be constructed based on the lengths of the Borders that are being shared between each pair of Regions (nodes). The process of calculating the length of the shared Borders between two Regions is presented in Alg. \ref{Alg:Algo1}. In case that two Regions are not neighbors then the corresponding matrix elements shall be equal to $0$. Furthermore, the diagonal elements of the adjacency matrix shall be equal to the length of the perimeter of the perspective Region. By doing so, we ensure that the ongoing traffic of a particular Region shall be the main factor in predicting its corresponding future state, while the rest of the Regions will influence the prediction results to a degree that is associated with the Borders that they share with the aforementioned Region. The Shared Borders Policy is presented in Alg. \ref{Alg:Algo2}. Upon the construction of the Adjacency Matrix $A$ using the Shared Borders Policy, it is required to normalize the resulting values in the $0-to-1$ range.

The main idea behind the incorporation of the Shared Borders Policy is that by enabling only the neighboring Regions to partake in the feature aggregation process, we are able to establish a distilled version of the spatial correlations that are inherent in topological structures. By doing so, we are alleviating part of the complexity that would otherwise be imposed on the forecasting model. Furthermore, by incorporating the aforementioned weights, we are able to establish a more refined encapsulation of the spatial dependencies that manifest.

\begin{algorithm}
\caption{LengthOfSharedBorder Algorithm}
\label{Alg:Algo1}
\begin{algorithmic}
\STATE \textbf{Description:} Algorithm for calculating the length of the overlapping section between two line segments in two-dimensional space.
\STATE \textbf{Input:} The endpoints of the two line segments: $(x_1, y_1), (x_2, y_2)$ for the first line segment and $(x_3, y_3), (x_4, y_4)$ for the second line segment.\\
\STATE \textbf{Output:} The $length$ of the overlapping segment between the two line segments.\\
\STATE \textbf{Begin algorithm}
\STATE 
\textbf{1.} If $x_1 > x_2$:\\
        $x_1, x_2 = x_2, x_1$\\
        $y_1, y_2 = y_2, y_1$\\
\textbf{2.} If $x_3 > x_4$:\\
        $x_3, x_4 = x_4, x_3$\\
        $y_3, y_4 = y_4, y_3$\\
\textbf{3.} Calculate the slopes of both line segments using the formulas: $slope_1 \gets \frac{y_2 - y_1}{x_2 - x_1}$ and $slope_2 \gets \frac{y_4 - y_3}{x_4 - x_3}$. If $slope_1=slope_2$ then the two line segments are parallel. If they are indeed parallel proceed to step 4, else $length=0$ and proceed to step 8.\\
\textbf{4.} If $\max(x_1,x_3) \leq \min(x_2,x_4)$, then the two line segments overlap. In that case, proceed to step 5, else $length=0$ and proceed to step 8.\\
\textbf{5.} Calculate $x_e$ and $x_s$, which are $x$ coordinates of the endpoints of the overlapping segment, by using the following formulas:\\
$x_s = \max(x_1,x_3)$\\
$x_e = \min(x_2,x_4)$\\
\textbf{6.} Calculate $y_e$ and $y_s$, which are $y$ coordinates of the endpoints of the overlapping segment, by using the following formulas:\\
$y_s = y_1 + \frac{(x_s - x_1) \cdot (y_2 - y_1)}{x_2 - x_1}$\\
$y_e = y_3 + \frac{(x_e - x_3) \cdot (y_4 - y_3)}{x_4 - x_3}$\\
\textbf{7.} Calculate the $length$ of the overlapping segment, by using the following formula: \\
$length = \sqrt{(x_e - x_s)^2 + (y_e - y_s)^2}$\\
\textbf{8.} Return $length$.\\
\STATE \textbf{End algorithm}

\end{algorithmic}
\end{algorithm}

\begin{algorithm}
\caption{Shared Borders Policy Algorithm.}
\label{Alg:Algo2}
\begin{algorithmic}
\STATE \textbf{Input:} The $N$ Regions, each represented as a list of 4-tuples. The size of each list of 4-tuples is equal to the number of edges that particular Region has. Each 4-tuple corresponds to the $x$ \& $y$ coordinates of the vertices that are connected by that particular edge.\\
\STATE \textbf{Output:} The weighted Adjacency Matrix $A$.\\
\STATE \textbf{Begin algorithm}
\STATE 
\textbf{1.} For each pair of lists (Regions) $i,j$:\\
\textbf{2.} \hspace*{3mm}Initialize $L\gets 0$ \\
\textbf{3.} \hspace*{3mm}For each pair of 4-tuples (edges) $k,l$:\\
\textbf{4.} \hspace*{6mm} $length \gets LengthOfSharedBorder(tuple_k,tuple_l)$\\
\textbf{5.} \hspace*{6mm} $L \gets L + length$\\
\textbf{6.} \hspace*{3mm} $A_i,j \gets L$\\
\textbf{6.} Return $A$\\
\STATE \textbf{End algorithm}

\end{algorithmic}
\end{algorithm}

\subsection{Adjustable Hops Policy}
The incorporation of a wide range of features is of paramount importance in the context of enabling the WEST GCN-LSTM model to conduct accurate predictions. In the previous subsection we showcased how the lengths of the shared Borders can be leveraged in order to construct the adjacency matrix $A$. In this subsection, we shall focus on showcasing how the Speed of a Population can be leveraged in order to calculate $K$, which refers to the number of stacked graph convolution layers incorporated at each constructed WEST GCN-LSTM model.

Let us consider three distinct parameters. The first one is the time between two consecutive predictions or observations and it corresponds to the selected time duration $t_{window}$. The second one is the average speed of a specific Population that is referred to as $Speed_p$. The third one is the average distance $D$ that has to be traversed in order to transit from the Center of one Region to the Center of one of its neighboring Regions. 

When the prediction process is being implemented on the basis of a singular Population, one is able to select the appropriate $t_{window}$ accordingly, in a manner that facilitates the establishment of consistent observations. For instance, in case that a Population moves at a significantly high speed, the chosen time between two consecutive observations can be decreased. Unfortunately, when a space is occupied by multiple Populations each one moving at a different Speed a problem arises, since in order to be able to formulate comparative analyses between them, it is necessary to utilize the same $t_{window}$ parameter across the various Populations. The WEST GCN-LSTM model, when leveraging the aforementioned Shared Borders Policy, receives as input only the Regional Traffic that corresponds to neighboring Regions. In other words, when using a conventional GCN-LSTM model the observed Populations should be able to traverse one Region at most during one time-step. As a result, in cases that a subset of the Populations is able to traverse multiple Regions within a singular time-step, a GCN-LSTM model that is using the Shared Borders Policy shall lose its advantage since there will be a potentially significant loss of information.

In order to mitigate this issue, we propose the Adjustable Hops Policy. According to the paradigm of GCNs, the number of graph convolution layers that are being deployed within a singular model corresponds to the number of aggregation hops that are being conducted each time the convolution process takes place. So in case of one convolution layer, only the first degree neighbors are taken into consideration, in case of two convolution layers, the first and second degree neighbors are taken into consideration, etc. 

The Adjustable Hops Policy commences by calculating the average distance between the Centers of the Regions and the identification of the Population with the lowest Speed. The parameter $t_{window}$ is then adjusted in a manner that this specific Population shall be capable of traversing at most one Region during each observation time-step of duration equal to $t_{window}$. This Population shall be used as the baseline and the dedicated GCN-LSTM model that corresponds to it shall have only one GCN layer, since its members shall be able to traverse a single Region at most during each time-step. By using the same $t_{window}$ parameter, calculate how many Regions at most each of the remaining Populations can traverse during a single time-stem. This number of Regions is equal to the number of GCN layers that shall be utilized for each of the corresponding Populations which is denoted by $K$. The specifics of the Adjustable Hops Policy are presented in Alg. \ref{Alg:Algo3}. It is worth mentioning that this policy results in the establishment of a dedicated prediction model for each distinct $K$ that emerges across the various Populations. Furthermore, in multi-step forecasting scenarios, like the ones that shall be explored in the next section of this work, $t_{window}$ should be adjusted in order to take into account the $U$ prediction steps that are being considered and thus needs to be multiplied by $\frac {U} {2}$.

\begin{algorithm}
\caption{Adjustable Hops Policy Algorithm.}
\label{Alg:Algo3}
\begin{algorithmic}
\STATE \textbf{Input:} The $Speed_p$ values, each one corresponding to one of the $P$ Populations. The $N$ 2-tuples, each one corresponding to the $x$ \& $y$ coordinates of a $Center_n$ of a Region. The $U$ value that corresponds to the number of prediction steps.\\
\STATE \textbf{Output:} The array $K$ of size $P$, each element of which corresponds to the maximum number of Regions, members of each corresponding Population can traverse during a singular time-step, whose duration is equal to $T$.\\
\STATE \textbf{Begin algorithm}
\STATE 
\textbf{1.} Initialize $T\gets 0$\\
\textbf{2.} Initialize $MinSpeed\gets Speed_0$\\
\textbf{3.} Initialize $D\gets 0$\\ 
\textbf{4.} For each pair of 2-tuples ($Center_n$) $k,l$:\\
\textbf{5.} \hspace*{3mm} $D = D + \sqrt{{(x_l - x_k)}^2 + {(y_l - y_k)}^2}$\\
\textbf{6.} $D = \frac{D}{N^2}$\\
\textbf{7.} For each $Speed$ value $p$:\\
\textbf{8.} \hspace*{3mm} If $Speed_p < MinSpeed$, then: $MinSpeed\gets Speed_p$\\
\textbf{9.} If $U > 1$: \\
\textbf{10.} \hspace*{3mm} $t_{window}\gets \frac{D}{Minspeed} \cdot \frac {U} {2}$\\
\textbf{11.} else:\\
\textbf{12.}\hspace*{3mm} $t_{window}\gets \frac{D}{Minspeed}$\\
\textbf{13.} For each $Speed$ value $p$:\\
\textbf{14.} \hspace*{3mm} $K_p = \text{round}\left(\frac{Speed_p \cdot t_{window}}{D}\right)$\\
\textbf{15.} Return $K$\\
\STATE \textbf{End algorithm}

\end{algorithmic}
\end{algorithm}

\section{Experimental Evaluation}
\label{Sec:experiment}
This section is dedicated to evaluating the efficiency of the proposed solution. Towards achieving this goal, the experimental evaluation consists of 19 forecasting models. The various forecasting models were designed and implemented via Python 3.9.13 and Tensorflow 2.9.1. Additionally, the Hardware Backend that was used for training and inference is a i5-11400 CPU and a NVIDIA GeForce RTX 3060 GPU. 

Furthermore, the experiments were conducted on the basis of a real and a synthetic dataset that was constructed using the Simulation of Urban MObility (SUMO) \cite{behrisch_sumo_2011} framework. SUMO is a versatile traffic simulator with the capability to handle extensive mobility networks. It incorporates various modes of transportation, including pedestrians, and is equipped with a plethora of tools for generating diverse mobility scenarios. The simulator exhibits realistic features of pedestrian mobility, such as pedestrian-pedestrian interactions in close proximity, reasonable walking speeds, and natural movement patterns. Additionally, SUMO enables pedestrians to interact safely by implementing features such as collision avoidance.

In order to maintain consistency throughout the experimental evaluation and across the two datasets, the authors of this work have chosen to implement the following format selections for both the experimental protocol and the parameter choices that supported the assessment of the models. The space that each dataset covers was divided into $6$ Regions. Each of the two datasets was split using the $80/20\%$ ratio for training and testing respectively. In the context of multi-step forecasting, $6$ time-steps were considered for the input and output sequences. Finally, the parameters of all of the examined forecasting models were tuned such as KerasTuner. 

\subsection{Evaluation Metrics}
The proposed model's performance is assessed using Mean Absolute Error (MAE), Mean Squared Error (MSE), and Root Mean Squared Error (RMSE), as evaluation metrics. These metrics are suitable for evaluating predictions of continuous numbers.

MAE measures the average magnitude of errors in a set of predictions as it is given in  Eq. \ref{eq:mae} where  $y_{i}$ are the ground truth values and $\hat{y_{i}}$ are the predictions, while $n$ is used to denote the total number of predictions in the evaluated/given set.  

\begin{equation} \label{eq:mae}
    MAE = \frac{1}{n} \sum_{i=1}^{n} \lvert y_{i} - \hat{y_{i}}\rvert 
\end{equation}

MSE is a commonly used metric for evaluating the average squared difference between observed values and predicted values in regression problems and is given in Eq. \ref{eq:mse}. 
\begin{equation}\label{eq:mse}
    MSE = \frac{1}{n} \sum_{i=1}^{n} (y_i - \hat{y}_i)^2
\end{equation}

Finally, RMSE is the squared version of MSE, it is used as a measure for the standard deviation of the prediction errors and it is given in Eq. \ref{eq:rmse}. 
\begin{equation} \label{eq:rmse}
    RMSE =  \sqrt{\frac{1}{n} \sum_{i=1}^{n} (y_{i} - \hat{y_{i}})^{2} }
\end{equation}

Using a combination of MAE, MSE, and RMSE provides a more comprehensive evaluation of a predictive model. Each metric captures different aspects of the model's performance, and using all three can offer a more nuanced understanding. MAE measures the average absolute errors, MSE measures the average squared errors, giving more weight to larger errors, and RMSE is the square root of MSE, providing an interpretable metric in the same unit as the data. The lower these metrics are, the more efficient the corresponding forecasting model is. 

\subsection{Benchmark Datasets}
In order to gain a comprehensive understanding of the proposed solution's capabilities, as well as of the experimental process that was employed in order to evaluate it, it is of paramount importance to delve deeper into the specifics of the two datasets that were used. These datasets are a subset of the Berlin (Cycling) dataset and the Central Park (Pedestrian) dataset, the latter of which was generated using SUMO. 

\begin{itemize}
\item \textit{Central Park}: As part of our study, we conducted a simulation of pedestrian traffic in the New York City area, over a period of seven days. Specifically, we focused on Central Park and the adjoining urban districts, modeling different traffic patterns. The generated dataset encompasses the movement attributes of 200,000-230,000 individuals on a daily basis, delineating their position and velocity per second. Central Park and the adjoining urban districts offer pedestrians the opportunity to take breaks and explore the different attractions or smart city features available to them, as well as to temporarily pause at points of interest such as interactive public art installations, historic landmarks, or food trucks and markets. Additionally, given the park's function as a sports venue, the simulation incorporates variable pedestrian speeds that simulate jogging or running. The distribution of speeds aims to realistically capture the motility characteristics of pedestrians based on factors including such as age, type of activity, and geospatial context. The $t_{window}$ for this dataset is equal to $5$ minutes.

\item \textit{Berlin}: The Berlin Cycling Dataset (\url{https://www.kaggle.com/datasets/phisinger/bike-counting-berlin}), procured by the Berlin administration in Berlin city, Germany, provides a comprehensive understanding of the long-term developments in bicycle traffic, including trends and seasonal fluctuations, through the use of Automatic Permanent Counting Points (APCPs). The dataset includes 9 years of bicycle traffic counts (2012-2020) collected using induction loops and sensors at APCPs. Bicycles passing over the detection cross-sections of the counting points are counted as they cause changes in the induced electromagnetic field, which are subsequently analyzed by the sensor and recorded as counting pulses. For roads with separate bicycle traffic guidance in each direction of travel, one counting station is established for each direction of travel, while for cross-sections with shared bicycle traffic guidance in both directions of travel, one counting station is installed for both directions of travel. The continuous counting process identifies bicycles based on specific geometries detected by the sensors while other vehicles are filtered out. The $t_{window}$ for this dataset is equal to $1$ hour.
\end{itemize}

Furthermore, the experimental evaluation process includes two additional datasets that are referred to as Central Park (Low) and Central Park (High). These two datasets derived from the Central Park dataset by categorizing the moving entities based on their Speed and assigning them to the corresponding dataset. More specifically, the Central Park (Low) consists of moving entities that can traverse utmost one Region during a single time-step ($K = 1$), while the Central Park (High) dataset consists of moving entities that can traverse utmost two Regions during a single time-step ($K = 2$). In other words, the aforementioned Central Park dataset is the amalgamation of the Central Park (Low) and Central Park (High) datasets. It is worth mentioning that these two datasets contain a similar total number of recorded moving entities (Central Park (High) contains about $4\%$ more moving entities compared to Central Park (Low)).

The aforementioned datasets can be leveraged to provide information regarding the Regions and the corresponding Regional Traffic that manifests during each time-step. In terms of Regional Traffic, both of these datasets enable direct access to the numerous recorded geolocation points throughout the duration of the simulation. However, the choice of an appropriate partitioning approach is contingent upon the specific context and challenges encountered. While these datasets provide valuable information regarding numerous recorded geolocation points, throughout the examined periods of time, they do not incorporate any notion of distinct spatial Regions. Thus, we had to construct the required Regions using the following approach. More specifically, we applied a k-means clustering algorithm to the various geolocation points that belong to each training dataset. The $k$ variable that corresponds to the k-means clustering algorithm was selected to be equal to $N=6$. The coordinates of the created centroids were selected to serve as the corresponding Centers of the $N$ Regions. Finally, dedicated Voronoi diagrams were created, for each dataset, using these $N$ Centers. The representation of the Voronoi diagram, for a set of $n$ points $P = \{(x_i, y_i)\}$, is achieved by using the following data structures:

\begin{itemize}
\item A list $V$ of Voronoi regions, where each region is associated with one of the input points.
\item A list $E$ of Voronoi edges that define the boundaries between regions.
\end{itemize}

\begin{figure*}[!h]
	\centering
		\includegraphics[scale=0.43]{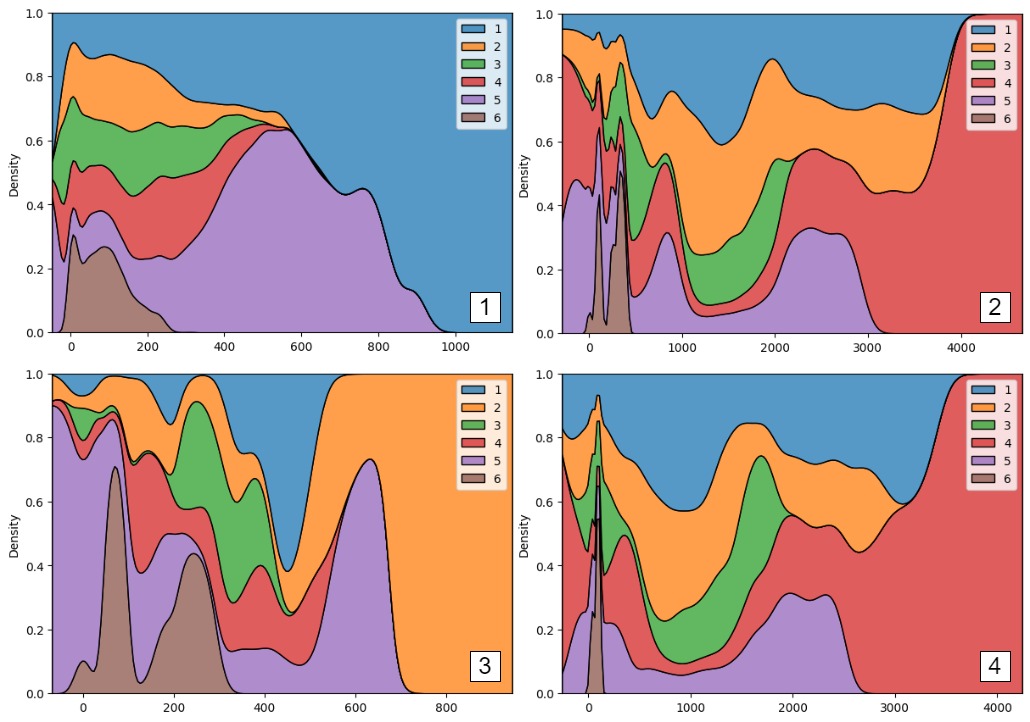}
	\caption{Densities of the regional traffic volumes that correspond to the: (1) Berlin , (2) Central Park, (3) Central Park (Low), and (4) Central Park (High) datasets.}
	\label{FIG:densities}
\end{figure*}

In order to calculate each Voronoi cell, one needs to find the bisectors between point $(x_i, y_i)$ and all other points in $P$, and then to clip the edges to the bounding box. Then, to merge any overlapping edges in $E$ to form the complete Voronoi diagram. This approach enables the examined Regions to encompass diverse geographic areas, exhibit varying population densities and feature distinct mobility patterns, while maintaining a constant size throughout the experimentation process. On top of that, the selection of datasets encompasses two distinct urban mobility scenarios (pedestrian and cycling), with the aim of embracing a broader spectrum of urban applications and assessing the models' capacity and generalization power across various vector input spaces that exhibit diverse statistical properties and characteristics. These properties and characteristics include different temporal granularities in the context of the chosen $t_{window}$, the size of the Regions, the trend, seasonality,and volume of regional traffic. For instance, Fig. \ref{FIG:densities} presents a comprehensive and nuanced portrayal of the different densities of the regional traffic volume that corresponds to each examined dataset using normalized Kernel Density Estimate (KDE) plots. KDE plots are a method for visualizing the distribution of observations in a dataset, by representing the data using a continuous probability density curve in a given number of dimensions. As the figure indicates there is a significant diversity in terms of regional traffic volume distribution across the various datasets and underlying Regions. This type of diversity of experimental conditions is of vital importance in the context of evaluating the efficiency of the proposed solution in a robust manner.

\subsection{Experimental Results}
Towards evaluating the efficiency of the proposed solution, on top of the aforementioned diversity of experimental conditions, the authors of this work have included a diverse ensemble of peer competitors. This ensemble consists of numerous forecasting models based on the LR ( LASSO \cite{haworth2014graphical}, Ridge \cite{kundu2020traffic}, Elastic Net \cite{liu2018short}, Lasso Lars \cite{gkountouna2020traffic}), the ML (KNN \cite{cai2020sample}, Decision Trees \cite{alajali2018intersection}, Tree Regression \cite{zhan2020multi}, Bagged Decision Trees \cite {xia2019traffic}, Random Forest \cite{evans2019forecasting}, Extra Trees Regressor \cite{mastelini2022online}), and the ED paradigms. Furthermore, it includes the three variations of the GCN-LSTM model that have been used in the context of regional traffic forecasting. The first one constructs the adjacency matrices based on the distance between the nodes of the graph, the second one creates the adjacency matrices on the basis of traffic pattern similarity matrices, and the third one establishes binary adjacency matrices based on whether or not the involved nodes are neighboring.

The latter of these variations also serves as part of an ablation study that investigates the performance of the proposed solution by removing certain components in order to understand the contribution of the component to the overall system. As stated before, alongside the rest of the  peer competitors, the proposed WEST GCN-LSTM is also tested against a GCN-LSTM model that uses binary graph convolution based on whether or not the examined Regions are neighboring. Furthermore, aside from the fully implemented WEST GCN-LSTM that leverages both proposed policies, we also explore a version of the WEST GCN-LSTM that is denoted as WE GCN-LSTM that solely relies on the Shared Borders Policy. The WE GCN-LSTM is suitable for Regional Traffic forecasting scenarios that do not incorporate the velocities of the various Populations, such as the Berlin dataset. Aside from the aforementioned dataset, we also considered a version of the Central Park dataset that does not include the velocities of the Populations. This ablation study serves as the third and final pillar that guarantees the robustness of the experimental evaluation process. 

In accordance with the aforementioned ablation study, the experiments are designed to evaluate the efficiency of the proposed solution in two distinct contexts: one that leverages both of the proposed policies (WEST GCN-LSTM) and one that uses only the Shared Borders Policy (WE GCN-LSTM). The evaluation process is conducted in the form of a comparative analysis against the various peer competitors. More specifically, Table \ref{we} displays the results of the WE GCN-LSTM model, compared against the various peer competitors in terms of MSE, RMSE, and MAE. These results correspond to the Berlin dataset, and Central Park dataset. For each of these datasets, the various forecasting models were trained and hyper-tuned independently. The displayed MSE, RMSE, and MAE values represent the average results (corresponding to the $6$ prediction time-steps and Regions) for each combination of forecasting models and datasets.

\begin{table*}[t]
 \caption{Experimental results for the use of the Shared Borders Policy (WE GCN-LSTM).}\label{we}
 \centering
    {\begin{tabular}{l c c c c c c} \hline
    \toprule
        \multirow{2}{*}{\textbf{Model}} & \multicolumn{3}{c}{\textbf{Berlin}} & \multicolumn{3}{c}{\textbf{Central Park}} \\ \cline{2-7} 
         & \textbf{MSE} & \textbf{RMSE} & \textbf{MAE} & \textbf{MSE} & \textbf{RMSE} & \textbf{MAE} \\ \hline
\textbf{LASSO} & 6813.74 & 82.576 & 55.856 & 28521.76 & 168.795  &  84.334 \\ 
\textbf{Ridge} & 6814.95 & 82.579 & 55.862 & 28550.85 & 168.974 &  84.462  \\ 
\textbf{Elastic Net} & 6813.86 & 82.576 & 55.855 & 28538.94 & 168.857 &  84.362  \\
\textbf{Lasso Lars} & 13969.78 & 118.304 & 91.777 & 19989.29 & 141.357 & 60.033  \\
\hline
\textbf{KNN} & 3040.22 & 55.111 & 33.134 & 148802.20 & 385.707 & 192.281  \\
\textbf{Decision Tree} & 5827.81 & 76.231 & 44.214 & 140607.77 & 375.077 & 181.793  \\
\textbf{Tree Regression} & 6746.09 & 82.151 & 47.549 & 143243.34 & 378.228 & 185.868  \\
\textbf{Bagged Decision Trees} & 3062.17 & 55.336 & 34.217 & 16767.66 & 409.477 & 194.014  \\
\textbf{Random Forest Reg} & 2742.04 & 52.215 & 34.337 & 198535.42 & 446.687 &  202.097\\
\textbf{Extra Trees Regressor} & 2934.09 & 54.107 & 35.935 & 206976.72 & 454.603 & 183.070\\
\hline
\textbf{LSTM ED} & 3678.18 & 60.565 &  42.507 & 145619.28 & 381.796 &  158.338\\
\textbf{BD-LSTM ED} & 2802.66 & 52.886 & 36.302 & 169972.55 & 412.567 & 199.281\\
\textbf{CNN-LSTM} & 3243.21 & 56.903 &  36.566 & 127684.06 & 355.516 &  123.458\\
\textbf{Hybrid LSTM ED} & 5063.64 & 71.291 & 49.423 & 49496.91 & 221.573 & 158.938\\
\textbf{Hybrid LSTM ATT ED} & 5872.85 & 76.589 & 51.105 & 59756.66 & 244.397 & 169.515\\
\hline 
\textbf{GCN-LSTM} (traffic) & 2515.62 & 50.156 & 31.744 & 14717.40 & 121.342 & 46.102\\
\textbf{GCN-LSTM} (centers) & 2914.56 & 53.969 & 33.730 & 21114.36 & 145.459 & 56.010\\
\textbf{GCN-LSTM} (binary) & 2332.97 & 48.263 & 30.354 & 16376.95 & 127.857 & 49.175\\ \hline \hline
\textbf{WE GCN-LSTM} (ours) & \textbf{1989.07} & \textbf{44.599} & \textbf{26.639} & \textbf{11718.71} & \textbf{108.253} & \textbf{42.594}\\ \hline        

\end{tabular}}
\end{table*}

Furthermore, Table \ref{west} displays the results of the WEST GCN-LSTM model, compared against the various peer competitors in terms of MSE, RMSE, and MAE. These results correspond to the Central Park (Low) \& (High) datasets. 

\begin{table*}[t]
\caption{Experimental results for the use of both policies (WEST GCN-LSTM). \label{west}}
 \centering
    {\begin{tabular}{l c c c c c c} \hline
    \toprule
        \multirow{3}{*}{\textbf{Model}} & \multicolumn{6}{c}{\textbf{Central Park}} \\ \cline{3-6} & \multicolumn{3}{c}{\textbf{Low}} & \multicolumn{3}{c}{\textbf{High}} \\ \cline{2-7} 
         & \textbf{MSE} & \textbf{RMSE} & \textbf{MAE} & \textbf{MSE} & \textbf{RMSE} & \textbf{MAE} \\ \hline
        
\textbf{LASSO} & 1957.84 & 44.275  & 27.170 & 38499.57 & 196.233  & 87.155 \\ 
\textbf{Ridge} & 1956.19 & 44.318 & 27.219 & 38694.22 & 196.735 & 87.312 \\ 
\textbf{Elastic Net} & 1956.68 & 44.289 & 27.187 & 38545.71 &  196.424 & 87.243 \\
\textbf{Lasso Lars} & 2288.14 & 47.848 & 34.168 & 32977.04 & 181.691 & 68.060 \\
\hline
\textbf{KNN} & 5153.61 & 71.738 & 47.242 & 211533.68 & 460.262 & 157.145 \\
\textbf{Decision Tree} & 7798.36 & 88.195 & 52.955 & 220273.80 & 468.947 & 171.820 \\
\textbf{Tree Regression} & 7892.61 & 88.865 & 55.491 & 205569.80 &  453.318 & 159.357  \\
\textbf{Bagged Decision Trees} & 4199.69 & 64.813 & 41.567 & 215214.93 & 464.097 & 168.351 \\
\textbf{Random Forest Reg} & 3947.21 & 62.900 & 39.293 & 216241.91 & 464.933 & 167.393 \\
\textbf{Extra Trees Regressor} & 3393.43 & 58.297 & 35.494 & 180371.66 & 424.022 & 152.694 \\
\hline
\textbf{LSTM ED} & 1401.66 & 37.422 &  27.564 & 63112.45 & 251.224 &  146.177 \\
\textbf{BD-LSTM ED} & 1440.24 & 37.943 & 27.442 & 57015.36 & 238.590 & 138.722 \\
\textbf{CNN-LSTM} & 1264.56 & 35.571 &  25.222 & 47348.78 & 217.745 &  122.168 \\
\textbf{Hybrid LSTM ED} & 1637.35 & 40.471 & 31.594 & 70813.06 & 266.018 & 165.205 \\
\textbf{Hybrid LSTM ATT ED} & 1705.97 & 41.316 & 31.202  & 93655.52 & 305.773 & 180.976 \\
\hline 
\textbf{GCN-LSTM} (traffic) & 1105.94 & 33.224 & 27.186 &  17656.01 & 132.887 & 57.776\\
\textbf{GCN-LSTM} (centers) & 1426.83 & 37.783 & 28.987 & 23634.92 & 153.532 & 73.110\\
\textbf{GCN-LSTM} (binary) & 1179.33 & 34.329 & 27.917 & 20463.79 &  142.843 & 63.485\\
\hline \hline
\textbf{WEST GCN-LSTM} (ours) & \textbf{802.87} & \textbf{28.335} & \textbf{22.861}  & \textbf{12106.60} & \textbf{110.030} & \textbf{47.904} \\ \hline 
\end{tabular}}

\end{table*}

\subsection{Discussion}
Before we proceed to analyzing the experimental results, it is of paramount importance to delve deeper into the intricacies of each dataset. As depicted in Fig. \ref{FIG:densities}, the regional traffic volume of the Berlin dataset is slightly larger compared to Central Park (Low). Furthermore, the volume of regional traffic of the Central Park and Central Park (High) datasets are quite similar, while both of them are significantly larger compared to the Regional Traffic (Low) dataset. This is due to the fact that in the context of the conducted experiments, fast-moving entities tend to emerge in a rather aperiodic manner, thus creating sudden bursts in the volume of regional traffic, contrary to slower-moving entities that are closely associated with periodic phenomena. Across all four datasets the larger regional traffic volume values are monopolized by a couple of regions. However, in the case of the Central Park and the Central Park (High) datasets, we witness that four regions are associated with quite large volumes of regional traffic. This inequality among Regions, in the context of their corresponding traffic, constitutes one of the most significant challenges that the various forecasting models have to overcome. More specifically, in the frame of the conducted experiments, we came across two types of regional traffic inequality, which are referred to as major inequality that corresponds to the Berlin and the Central Park (Low) datasets, and as minor inequality that corresponds to the Central Park and the Central Park (High) datasets. The manifestation of each type of regional traffic inequality highly depends on the moving speed of the various entities.

The experimental results indicate a plethora of interesting insights that are worth exploring. Let us begin our exploratory analysis by not considering the forecasting models that are based on the GCN-LSTM paradigm. When doing so, we see that for the Berlin dataset ML and ED solutions manage to perform the best, for the Central Park and Central Park (High) datasets LR solutions manage to outperform their competitors, and for the Central Park (Low) dataset ED solutions achieve the best scores. In other words, the more advanced ED solutions are better equipped to handle scenarios of major regional traffic inequality, while the quite simplistic LR solutions manage to outperform their competitors in scenarios that are characterized by minor regional traffic inequality. This is due to the fact that LR models are less sensitive to input fluctuations and as such, they tend to converge to none optimal solutions. Thus, in the case of minor regional traffic inequality, the extent that they take into account the couple of Regions that monopolize traffic volume does not alter significantly the basis upon which they conduct their predictions, and consequently produce superior results.

When we include forecasting models that are based on the GCN-LSTM paradigm, we see that these models manage to either outperform or perform very closely to the best of their competitors across all examined datasets. This showcases the fact that the information filtering process that is intertwined with graph convolution is capable of mitigating the rather negative impact that regional traffic inequalities have on the various forecasting models. However, the relative performance of each of these models, when compared to its peers, is closely associated with the reasoning behind each information distillation process and the characteristics of each dataset. The performance of the GCN-LSTM (centers) model highly depends on the topology of the Regions and the assumption that the optimal way to represent the relation between two Regions is the proximity to each other. Since in our experiments we did not consider homogeneously shaped Regions, this approach produced the worst results in terms of prediction accuracy compared to its peers. Nevertheless, it produced results that are similar to the best LR, ML, and ED models. On the contrary, the performance of the GCN-LSTM (traffic) model depends on the traffic correlations among the various Regions. Real data exhibit a wide range of statistical properties within entity motion patterns. Consequently, the traffic-based relations among Regions that the  GCN-LSTM (traffic) is designed to exploit are more complex and harder to encapsulate in the case of real data. As a result, the GCN-LSTM (traffic) model managed to produce good results in the case of the Central Park, Central Park (Low), and Central Park (High) datasets, while it performed less optimally in the case of the Berlin dataset. Last but not least, the GCN-LSTM (binary) model managed to perform, relatively to its peers, in the most consistent manner across all datasets. In fact, it managed to outperform the GCN-LSTM (centers) model across all datasets, and the GCN-LSTM (centers) in the case of the Berlin dataset. This serves as a quite significant indication that the choice to construct the adjacency matrices based on the neighboring status of the various Regions is a robust approach that manages to provide satisfying results, regardless of the characteristics of each examined dataset. 

\begin{figure*}
	\centering
		\includegraphics[scale=0.7]{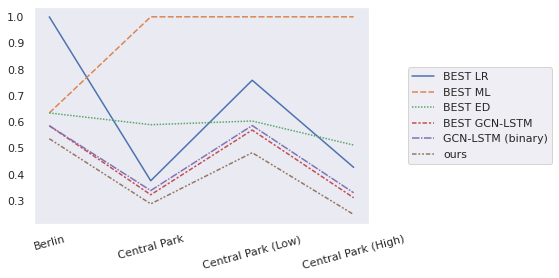}
	\caption{Normalized RMSE values that correspond to the proposed solution (ours), to GCN-LSTM (binary), as well as to the best LR, ML, ED, and GCN-LSTM solutions.}
	\label{FIG:norm}
\end{figure*}

As stated before, the proposed solution is an attempt at refining the aforementioned neighbor-based approach. So let us examine the performance of the proposed solution using Fig. \ref{FIG:norm}. Fig. \ref{FIG:norm} displays the performance of the proposed solution, of the best LR, ML, ED, and GCN-LSTM solutions, as well as of the GCN-LSTM (binary) model, in a comparative manner that is based on the use of normalized RMSE values. The same process was also performed for the MSE and the MAE results, and despite some minor differences, the overall conclusions remained the same. Thus, we decided to not include them as part of this discussion. The proposed solution clearly outperformed all of its competitors, across all examined datasets. More specifically, for the Berlin dataset, the proposed solution outperformed the best LR model by $46\%$, the best ML model by $15\%$, the best ED model by $15\%$, and the best GCN-LSTM model by $8\%$. Furthermore, for the Central Park dataset, the proposed solution outperformed the best LR model by $23\%$, the best ML model by $71\%$, the best ED model by $51\%$, and the best GCN-LSTM model by $10\%$. For the Central Park (Low) dataset, the proposed solution outperformed the best LR model by $36\%$, the best ML model by $51\%$, the best ED model by $20\%$, and the best GCN-LSTM model by $15\%$. Finally for the Central Park (High) dataset, the proposed solution outperformed the best LR model by $39\%$, the best ML model by $73\%$, the best ED model by $49\%$, and the best GCN-LSTM model by $16\%$.

The aforementioned experimental results are indicative of the fact that the proposed solution is capable of outperforming its competitors by a quite significant margin, in a consistent manner, across all examined datasets. However, it is of paramount importance to also examine the proposed solution in a manner that is aligned with the ablation study paradigm, in order to showcase the effect of each part of the proposed solution on its overall performance. The proposed solution outperformed the GCN-LSTM (binary) model by $8\%$ for the Berlin dataset, by $14\%$ for the Central Park dataset, by $17\%$ for the Central Park (Low) dataset, and by $22\%$ for the Central Park (High) dataset. As mentioned earlier, the Berlin and Central Park datasets were used to test a version of the proposed solution that uses only the Shared Borders Policy (WE GCN-LSTM). Thus, the fact that the proposed solution that leverages both policies (WEST GCN-LSTM) outperformed the GCN-LSTM (binary) by a greater margin, when compared to the margin by which WE GCN-LSTM surpassed GCN-LSTM (binary), showcases that the fully implemented proposed solution that uses both policies is more efficient. The same principle applies when comparing the proposed solution against the rest of its competitors. The WE GCN-LSTM outperformed the second-best model by $8\%$ for the Berlin dataset, and by $10\%$ for the Central Park dataset, while the WEST GCN-LSTM model outperformed the second-best model by $15\%$ for the Central Park (Low) dataset, and by $16\%$ for the Central Park (High) dataset. Since the Central Park dataset is the amalgamation of the Central Park (Low) and the Central Park (High) datasets, experimental consistency, in regards to the comparison between the WE GCN-LSTM and the WEST GCN-LSTM models, is achieved via the use of combinations of datasets that entail identical regional traffic characteristics, as well as across datasets that exhibit different types of regional and traffic-based characteristics.

\section{Conclusions and Future Research}
\label{Sec:Conclusion}
The objective of this work is to introduce an advanced prediction model aimed at enhancing the accuracy of regional traffic forecasting through the refinement and fusion of information. Towards achieving this goal, we extend spatio-temporal graph neural networks in alignment with the regional traffic forecasting paradigm. Our approach is based on the proposal of a novel architecture for spatio-temporal graph neural networks, incorporating weighted stacked graph convolutions. The implementation of weighted stacked convolutions necessitates the calculation of a weighted adjacency matrix (denoted as $A$) and determining the number of graph convolution layers (denoted as $K$). To address this, we introduce two novel algorithms that utilize information about the speed of the populations and the topography of the regions that they traverse to compute the adjacency matrix and $K$. In the context of this work, these policies are referred to as the Shared Borders Policy and the Adjustable Hops Policy. This implementation design enables the proposed solution to encapsulate both the spatial and temporal characteristics that are intertwined with regional traffic forecasting in an optimal and more refined manner.

In order to evaluate the efficiency of the proposed solution, we conducted numerous experiments. The proposed solution managed to significantly outperform its competitors in the frame of an experimental evaluation that consists of 19 forecasting models, across all examined datasets. This is due to the fact that the proposed solution manages to simultaneously encapsulate both temporal and spatial aspects of regional traffic forecasting, through information fusion. Furthermore, through information distillation, the proposed solution is also capable of mitigating the dire ramifications of regional traffic inequality. Finally, an additional ablation study concluded that each of one the three parts of the proposed solution serves towards boosting the performance of the proposed solution. 

In terms of future research, there are are various directions in order to potentially enhance the performance of the proposed solution even further. One of these directions derives from the fact that the experimental evaluation process did not account for various forms of obstacles that may affect the resulting weighted Adjacency Matrix $A$ that is calculated on the basis of the Shared Borders Policy. Thus, a more refined version of Alg. \ref{Alg:Algo1} that takes into consideration such obstacles could result in a more efficient prediction model.
Furthermore, it is worth examining the use of hybrid approaches for constructing the weighted Adjacency Matrix $A$. These hybrid approaches could stem from the combination of several methodologies that have been already examined in the frame of the corresponding scientific literature, and the proposed solution.

\section*{Acknowledgment}

This project has received funding from the European Union’s Horizon 2020 research and innovation programmes under grant agreements No 101016509 (CHARITY) and No 777695 (MASTER). The work reflects only the authors’ view, and the EU Agency is not responsible for any use that may be made of the information it contains.

\bibliographystyle{IEEEtran}
\bibliography{references}

\end{document}